%% file: main.tex
\documentclass[openacc]{rstransa}

\usepackage{relsize}
\usepackage{mathtools} 
\usepackage{booktabs}
\usepackage{multirow}
\usepackage{gensymb}

\begin{document}

\title{Unraveled Multilevel Transformation Networks for Predicting Sparsely-Observed Spatiotemporal Dynamics}

\author{
Priyabrata Saha and Saibal Mukhopadhyay}

\address{School of Electrical and Computer Engineering, Georgia Institute of Technology, Atlanta, GA 30332, USA}

\subject{xxxxx, xxxxx, xxxx}


\corres{Priyabrata Saha\\
\email{priyabratasaha@gatech.edu}}

\begin{abstract}
In this paper, we address the problem of predicting complex, nonlinear spatiotemporal dynamics when available data is recorded at irregularly-spaced sparse spatial locations. 
Most of the existing deep learning models for modeling spatiotemporal dynamics are either designed for data in a regular grid or struggle to uncover the spatial relations from sparse and irregularly-spaced data sites. We propose a deep learning model that learns to predict unknown spatiotemporal dynamics using data from sparsely-distributed data sites. We base our approach on Radial Basis Function (RBF) collocation method which is often used for meshfree solution of partial differential equations (PDEs). The RBF framework allows us to unravel the observed spatiotemporal function and learn the spatial interactions among data sites on the RBF-space. The learned spatial features are then used to compose multilevel transformations of the raw observations and predict its evolution in future time steps. 
We demonstrate the advantage of our approach using both synthetic and real-world climate data.       
\end{abstract}


\begin{fmtext}
\section{Introduction}
Many real-world processes, for example, climate and ocean dynamics \cite{marshall1989atmosphere}, epidemic dynamics \cite{hethcote2000mathematics}, and neurological signals \cite{izhikevich2007dynamical} to name a few, are spatiotemporal in nature \cite{cressie2015statistics, osama2018learning}. Generally, physical processes are modeled by systems of nonlinear spatiotemporal differential equations constructed based on physical laws and elaborate experiments. However, for many practical problems, the spatiotemporal interactions within the system are highly nonlinear and complex to describe analytically \cite{ayed2019learning, saha2021deep}. Recently, machine learning methods, particularly deep learning, have shown promise in auto- 
\end{fmtext}


\maketitle
\input{introduction}
\input{background}

\input{method}
\input{expSetup}

\input{results}
\input{conclusion}


\bibliographystyle{rsta}
\bibliography{ref}

\end{document}

%% file: introduction.tex
\begin{figure}[b]
\centering\includegraphics[width=1\linewidth]{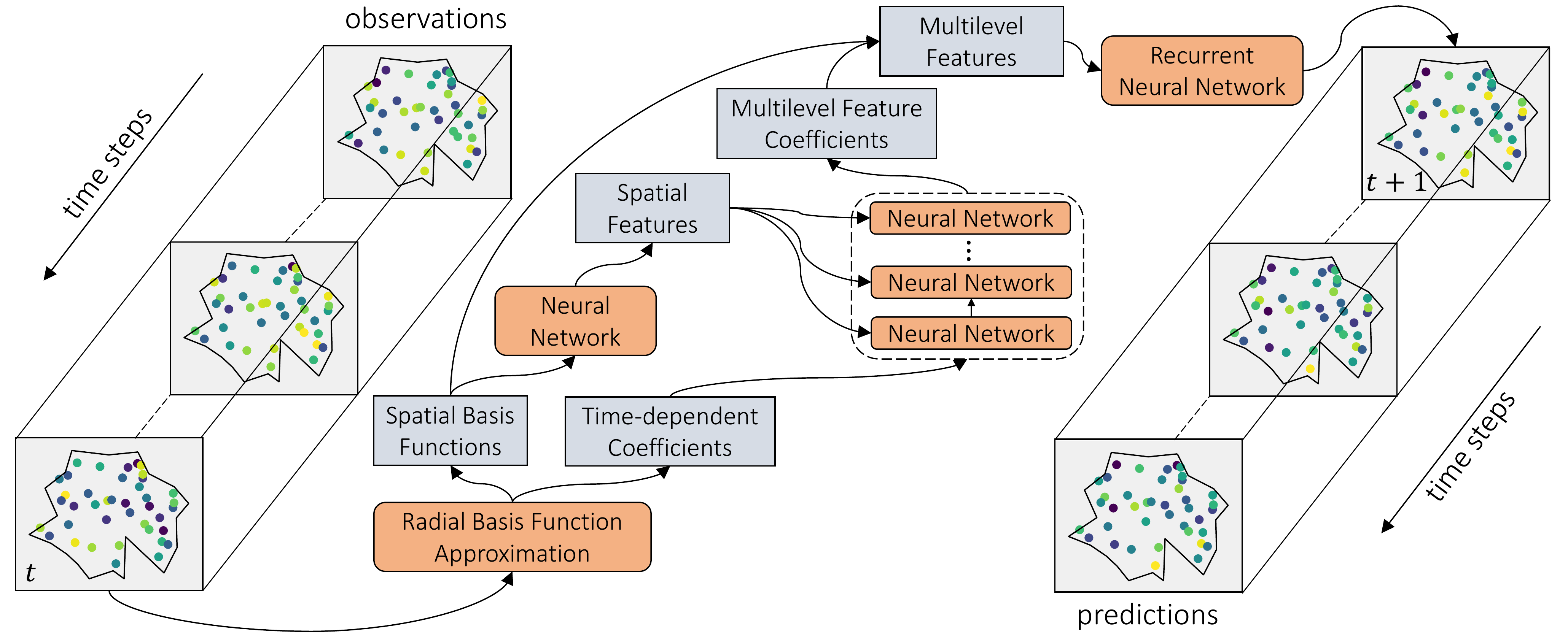}
\caption{Outline of the proposed approach. $t$ denotes the current time step. Latent spatial transformations are learned on the RBF-space, using a neural network. The learned spatial features are then integrated with the RBF coefficients through a cascade of neural networks to compose multilevel features. These multilevel features are finally looped through a recurrent network across time steps to learn the temporal relationships and predict the future observations.}
\label{fig:intro}
\end{figure}

\hspace{-5.3mm} matically learning the spatiotemporal relations embedded within the observations
\cite{xingjian2015convolutional, long2018pde, de2019deep, rudy2017data}. Despite promising results, 
deep learning methods usually require large amount of data for generalized solutions and pose many challenging scenarios when only a limited number of observations are available. In many real-world scenarios, collecting a large amount of data is very challenging if not impossible. One such challenge is that sensors for measurement or observations can only be placed at a few scattered locations, which is a very common scenario for real-world systems. For examples, distributions of weather stations for meteorological data collection vary from region to region.    

Most of the existing deep learning methods assume either densely sampled observations are available for learning or physical equations of the underlying system is known apriori. For example, a class of methods \cite{raissi2019physics, sirignano2018dgm, raissi2018deep} model the solution of a PDE as function of spatial and temporal coordinates and incorporate the known PDE in the loss function by evaluating the gradients of the solution. These methods do not generalize well with limited data when the underlying PDE is unknown.  

On the other side of the spectrum, Long et al. \cite{long2018pde}, Ruthotto and Haber \cite{ruthotto2020deep} proposed new convolutional neural networks (CNNs) involving constrained kernels to uncover unknown PDEs. However, these methods are designed assuming that data is available in a dense regular grid. Some CNN-based approaches also incorporate the prior knowledge of the physical equations in model architecture for various purposes. de Bezenac et al. \cite{de2019deep} assimilated the general solution of the advection-diffusion equation into a CNN architecture for sea surface temperature prediction. Physical models and deep networks are integrated together in \cite{long2018hybridnet, saha2021physics, guen2020augmenting} to learn dynamics where the given physical models are inadequate to describe the observed dynamics. Regular grid convolutional operations make these methods unsuitable for modeling with sparse irregular data. 

Some recent work address the problem of learning spatiotemporal dynamics from scattered data using graph networks (GNs) \cite{belbute2020combining, iakovlev2020learning, seo2019differentiable, seo2019physics}. Belbute-Peres et al. \cite{belbute2020combining} augmented a low-resolution CFD solver with GN to learn high-resolution solution. Their method is suitable for solving known PDEs.  Iakovlev et al. \cite{iakovlev2020learning} proposed a GN method to learn continuous-time PDEs from sparse data. The continuous-time formulation assumes the temporal evolution of the systems only depends on the current state and its spatial derivatives, which may not be valid for real-world system.
Seo and Liu \cite{seo2019differentiable} proposed a differentiable
physics-informed graph network (DPGN) incorporating data-specific physical equations into GNs. A more generic spatial difference layer on GN is proposed in \cite{seo2019physics}, which can learn unknown spatiotemporal dynamics. 

In this paper, we consider a different approach for modeling spatiotemporal dynamics from sparsely-observed data, which is inspired by the RBF collocation method for solving PDEs. Our goal is to develop a model that can learn the spatiotemporal interactions from a sequential observations of length $\tau$ at irregularly-spaced $n$ data sites and predict a length-$T$ sequence of future observations at those data sites. A more formal definition of the problem is given in section \ref{method}. Similar to the RBF collocation method, we assume the observation function (at each time step) is a linear combination of a set of RBFs centered at the data sites. This formulation allows us to learn the unknown spatial transformations separately using a deep neural network on the shared RBF-space (instead of on the raw observation variables). 

Integration of deep learning with RBF collocation method was first proposed in our prior work \cite{saha2021deep} to model unknown PDEs from scattered observations. The method presented in \cite{saha2021deep} is designed assuming the underlying dynamical system follows a time-dependent nonlinear PDE. It applies spatial transformation only on the raw observations. However, many dynamical systems, for example, certain nonlinear diffusion, involve spatial differential operations not only on the direct observation variable but also on its higher level nonlinear transformations. Furthermore, a fixed temporal relation between successive observations is applied in \cite{saha2021deep}, assuming a good estimate of the temporal order of the underlying PDE. Many real-world dynamical systems can involve arbitrarily complex spatiotemporal relations and the aforementioned assumptions may not be appropriate. In this paper, we introduce a deep learning model that addresses the forenamed limitations and is suitable for generic spatiotemporal dynamical systems which may or may not follow a PDE.   

We propose to address the first problem by applying the spatial transformations to the latent feature coefficients at multiple levels, obtained from the RBF coefficients of the raw observations through a cascade of neural networks. The second problem is handled using a recurrent network that fuses the multilevel features across time steps to predict the future observations. A graphic outline of the proposed approach is shown in Figure \ref{fig:intro}. We evaluate our method in the forecasting task for synthetic as well as real-world spatiotemporal data. Experiments show that our method outperforms the baselines and also uses significantly less number of parameters compared to the best performing baseline. 

%% file: background.tex
\section{RBF Collocation Method}
\label{rbf_collocation}

In this section, we briefly introduce the method of RBF collocation for solving spatiotemporal dynamical systems. Consider the following problem.

\textit{Find a (continuous) function $u: \Omega \times [0, t_{end}] \rightarrow \mathbb{R}$ that satisfies the following state equation
\begin{equation}
    \frac{\partial u}{\partial t} = Lu, \quad \mathbf{x} \in \Omega \subset \mathbb{R}^d, \quad 0 \leq t \leq t_{end},
    \label{eqn:linear_pde}
\end{equation}}
\textit{ \quad
and initial condition
\begin{equation}
    u(0, \mathbf{x}) = w(\mathbf{x}), 
    \quad \mathbf{x} \in \Omega
    \label{eqn:linear_pde_ic}
\end{equation}
}
Here $L: \mathbb{R} \rightarrow \mathbb{R}$ is some linear (spatial) differential operator. (\ref{eqn:linear_pde}) essentially describes a time-dependent linear \textit{partial differential equation} (PDE). 

In RBF collocation method, one considers a set of $n$ collocation nodes $\mathcal{X} = \{\mathbf{x}_1, \mathbf{x}_2, \cdots, \mathbf{x}_n\} \subset \Omega$ and assumes the solution function $u(t, \mathbf{x})$ to be a linear combination of a set of RBFs centered at those collocation nodes, i.e.,
\begin{equation}
     u(t, \mathbf{x}) = \sum_{j=1}^n c_j (t) \varphi(\|\mathbf{x} - \mathbf{x}_j\|), \quad \mathbf{x} \in \Omega, \quad t \in [0, t_{end}]
    \label{eqn:linear_pde_radial_approximation}
\end{equation}
Here $\|\cdot\|$ denotes some norm in $\mathbb{R}^d$, usually the Euclidean norm and $\varphi$ is a \textit{strictly positive definite radial function}. $c_j(t), \ j = 1, 2, \cdots, n$ are the unknown time-dependent coefficients to be determined at each time step whereas the spatial function $\varphi$ is independent of time. The space-time separated formulation and the linearity of the spatial operator $L$ allow the spatial and temporal operations to be independent of each-other.

Inserting (\ref{eqn:linear_pde_radial_approximation}) into (\ref{eqn:linear_pde}) and (\ref{eqn:linear_pde_ic}) we get the following \textit{ordinary differential equation} (ODE) 
\begin{align}
    \sum_{j=1}^n \frac{dc_j(t)}{dt}  \varphi(\|\mathbf{x} - \mathbf{x}_j\|) - c_j(t) L\varphi(\|\mathbf{x} - \mathbf{x}_j\|) = 0,  \quad \mathbf{x} \in \Omega, \quad 0 \leq t \leq t_{end}, 
    \label{eqn:ode}
\end{align}
with initial condition
\begin{equation}
    \sum_{j=1}^n c_j (0) \varphi(\|\mathbf{x} - \mathbf{x}_j\|) = w(\mathbf{x}), \quad \mathbf{x} \in \Omega
    \label{eqn:ode_ic}
\end{equation}

Approximating the time-derivatives $\frac{dc_j(t)}{dt}$ by first order finite difference, we get
\begin{align}
     \sum_{j=1}^n  c_j(t + \Delta t) \varphi(\|\mathbf{x} - \mathbf{x}_j\|)  = \sum_{j=1}^n c_j(t) \Big( \varphi(\|\mathbf{x} - \mathbf{x}_j\|) + \Delta t L\varphi(\|\mathbf{x} - \mathbf{x}_j\|)\Big), \quad \mathbf{x} \in \Omega, \quad 0 \leq t < t_{end},
    \label{eqn:ode_fd}
\end{align}
where $\Delta t > 0$ is the step-size in time. Applying (\ref{eqn:ode_ic}) and (\ref{eqn:ode_fd}) on the set of collocation nodes $\mathcal{X}$, we get the following discrete-time linear dynamical system. 
\begin{align}
    \Phi \mathbf{c}_{t+\Delta t} &= H\mathbf{c}_t,  \quad  t \in [0, \Delta t, 2 \Delta t, \cdots, t_{end}-\Delta t] \nonumber \\ \Phi \mathbf{c}_0 &= \mathbf{w}
    \label{eqn:compact}
\end{align}
Here $\Phi$ is the RBF interpolation matrix whose entries are given by $\Phi_{ij} = \phi(\|\mathbf{x}_i - \mathbf{x}_j\|)$, $\  i,j = 1,2, \cdots, n$ and $\mathbf{c}_t = [c_1(t), c_2(t),\cdots, c_n(t)]^\top$ is the coefficient vector at time $t$. The elements of the matrix $H$ are given by
\begin{align}
    H_{ij} = \varphi(\|\mathbf{x}_i - \mathbf{x}_j\|) + \Delta t L\varphi(\|\mathbf{x} - \mathbf{x}_j\|)|_{\mathbf{x} = \mathbf{x}_i}
\end{align}
$\mathbf{w} = [w(\mathbf{x}_1), w(\mathbf{x}_2), \cdots, w(\mathbf{x}_n)]^\top$ is the vector of initial values at the collocation nodes.
Given the existence of the inverse of the interpolating matrix $\Phi$, the coefficient vectors $\mathbf{c}_t$ are uniquely determined by iteratively solving (\ref{eqn:compact}). Obtained coefficient vectors $\mathbf{c}_t$ are then used to compute the numerical solution $u(t, \mathbf{x})$ by (\ref{eqn:linear_pde_radial_approximation}). For a large class of radial functions including (inverse) multiquadrics, Gaussian, it has been proven in the literature that the matrix $\Phi$ is non-singular (and therefore, \textit{invertible}) if the data sites are all distinct \cite{micchelli1984interpolation,broomhead1988multivariable, fasshauer2007meshfree}. Note, we have ignored the boundary condition of the given PDE for brevity, which can be included on the right hand side of (\ref{eqn:compact}). Further details can be found in \cite{fasshauer2007meshfree}.

%% file: method.tex
\section{Unraveled Multilevel Transformation Network (UMTN)}
\label{method}

In this paper, we study the following problem of modeling spatiotemporal dynamics for multi-step prediction from scattered (and sparse) observation. 

\textbf{Problem.} \textit{
Consider an example sequence $(\mathcal{X}, \mathbf{u}_t), \ t = 0, 1, \cdots, (T+\tau-1)$ from a spatiotemporal dynamical system, where $\mathcal{X} = \{\mathbf{x}_1, \mathbf{x}_2, \cdots, \mathbf{x}_n\} \subset \Omega \subset \mathbb{R}^d$ being the set of data (measurement) sites and $\mathbf{u}_t = [u(t, \mathbf{x}_1), u(t, \mathbf{x}_2), \cdots, u(t, \mathbf{x}_n)]^\top$ being the corresponding data (measurement) values at time step $t$, with $u(t,\mathbf{x}_i) \in \mathbb{R}$. Given such $N$ examples $\big(\mathcal{X}^{(k)}, \mathbf{u}_t^{(k)}\big), \ t = 0, 1, \cdots, (T+\tau-1), \ k = 1, 2, \cdots, N$, find a (nonlinear) function $F:\mathbb{R}^{n \times d} \times \mathbb{R}^{n \times \tau} \rightarrow \mathbb{R}^{n \times T}$, \  $\tau > 0, T > 0$ that satisfies the conditions 
\begin{equation}
  \mathbf{u}_{T+\tau-1}^{(k)}, \cdots, \mathbf{u}_{\tau+1}^{(k)}, \mathbf{u}_{\tau}^{(k)} = F\big(\mathcal{X}^{(k)}, \mathbf{u}_{\tau-1}^{(k)}, \cdots, \mathbf{u}_1^{(k)}, \mathbf{u}_0^{(k)}\big),  \quad k = 1,2, \cdots, N
\end{equation}
}

We consider representing the function $F$ as a deep neural network $F_{\theta}$, with parameter vector $\theta$. Here we propose an architecture for $F_\theta$ that is capable of modeling the spatiotemporal interactions of the underlying system. This point onward, we describe the process for a single example sequence $(\mathcal{X}, \mathbf{u}_t), \ t = 0, 1, \cdots, (T+\tau-1)$ and omit the example specific superscript notation $\cdot^{(k)}$ for brevity.

We base our model on RBF collocation method for solving time-dependent linear PDEs, described in section
\ref{rbf_collocation}. Likewise, we consider the observed spatiotemporal function $u(t, \mathbf{x})$ to be a linear combination of a set of RBFs centered at the data sites $\mathbf{x}_j$, i.e.,
\begin{equation}
     u(t, \mathbf{x}) = \sum_{j=1}^n c_j (t) \varphi(\|\mathbf{x} - \mathbf{x}_j\|), \quad \mathbf{x} \in \Omega, \quad t = 0, 1, \cdots, (T+\tau-1)
     \label{eqn:obsd_func_rbf_apprx}
\end{equation}
Applying (\ref{eqn:obsd_func_rbf_apprx}) at the data sites of $\mathcal{X}$ we get following matrix-vector compact form,
\begin{equation}
    \mathbf{u}_t = \Phi \mathbf{c}_t
    \label{eqn:compact_linear_sys}
\end{equation}
where $\Phi$ is RBF matrix and $\mathbf{c}_t$ is the coefficient vector, as defined in section 
\ref{rbf_collocation}.
We assume that the spatial component of the observed function $u$ remains same across all time steps and all intermediate transformations.  Unraveling space from the data using RBFs allows learning the spatial interaction among data sites and then applying the learned transformation to the coefficients $c_j$. We first discuss how the RBF collocation method of section 
\ref{rbf_collocation} can be augmented with deep learning to learn an unknown first order linear spatiotemporal dynamical system from scattered observation. Subsequently, we propose how the model for the linear system can be used as a building block to design deep neural networks for more general spatiotemporal dynamical systems.    
\subsection{Linear Spatial Transformation Block (LSTB)}
\label{lstb}

We use the method proposed in \cite{saha2021deep} to model an unknown first order linear spatiotemporal dynamical system from scattered observation. An example system can be described by (\ref{eqn:linear_pde}), where the linear (spatial) differential operator $L$ is unknown. Since we assume an RBF-based representation of the observed function $u$, we can learn the application of unknown spatial operation $L$ on the chosen RBF $\varphi$, instead of the actual observed function $u$. Given any two data sites $\mathbf{x}_i, \mathbf{x}_j$, and the corresponding RBF value $\varphi(\|\mathbf{x} - \mathbf{x}_j\|)$, we use a neural network $S_\alpha$ to approximate $L\varphi(\|\mathbf{x} - \mathbf{x}_j\|)\big|_{\mathbf{x}=\mathbf{x}_i}$, i.e.,
\begin{equation}
    \Delta t L\varphi(\|\mathbf{x} - \mathbf{x}_j\|)\big|_{\mathbf{x}=\mathbf{x}_i} = S_\alpha \big(\mathbf{x}_i, \mathbf{x}_j, \varphi(\|\mathbf{x}_i - \mathbf{x}_j\|)\big)
\end{equation}
Here $\alpha$ denotes the parameter vector of the neural network $S_\alpha$. Note, we are specifying the neural network $S_\alpha$ to learn the time difference $\Delta t$ between successive measurements and hence, eliminating the $\Delta t$ notation from this point onward. 
Accordingly, the discrete time linear dynamical system of (\ref{eqn:compact}) can be rewritten as 
\begin{equation}
    \Phi \mathbf{c}_{t+1} = H\mathbf{c}_t = (\Phi + [S_\alpha]) \mathbf{c}_t,  \quad  t = 0, 1, \cdots (T+\tau-2),
    \label{eqn:compact_linear_dynamics}
\end{equation}
where $[S_\alpha]$ is the RBF spatial transformation matrix, i.e.,  
\begin{equation}
    [S_\alpha]_{ij} = S_\alpha \big(\mathbf{x}_i, \mathbf{x}_j, \varphi(\|\mathbf{x}_i - \mathbf{x}_j\|)
\end{equation}
Applying (\ref{eqn:compact_linear_sys}) into (\ref{eqn:compact_linear_dynamics}), we can directly get the next step prediction:
\begin{equation}
    \widehat{\mathbf{u}}_{t+1} = \mathbf{u}_t + [S_\alpha] \Phi^{-1} \mathbf{u}_t, \quad  t = 0, 1, \cdots (T+\tau-2)
\end{equation}
Note, we use $\widehat{\mathbf{u}}_t$ to denote prediction while $\mathbf{u}_t$ denotes the ground truth. More details regarding this method of modeling unknown linear spatiotemporal dynamics can be found in \cite{saha2021deep}. 

\begin{figure}[t]
\centering\includegraphics[width=0.85\linewidth]{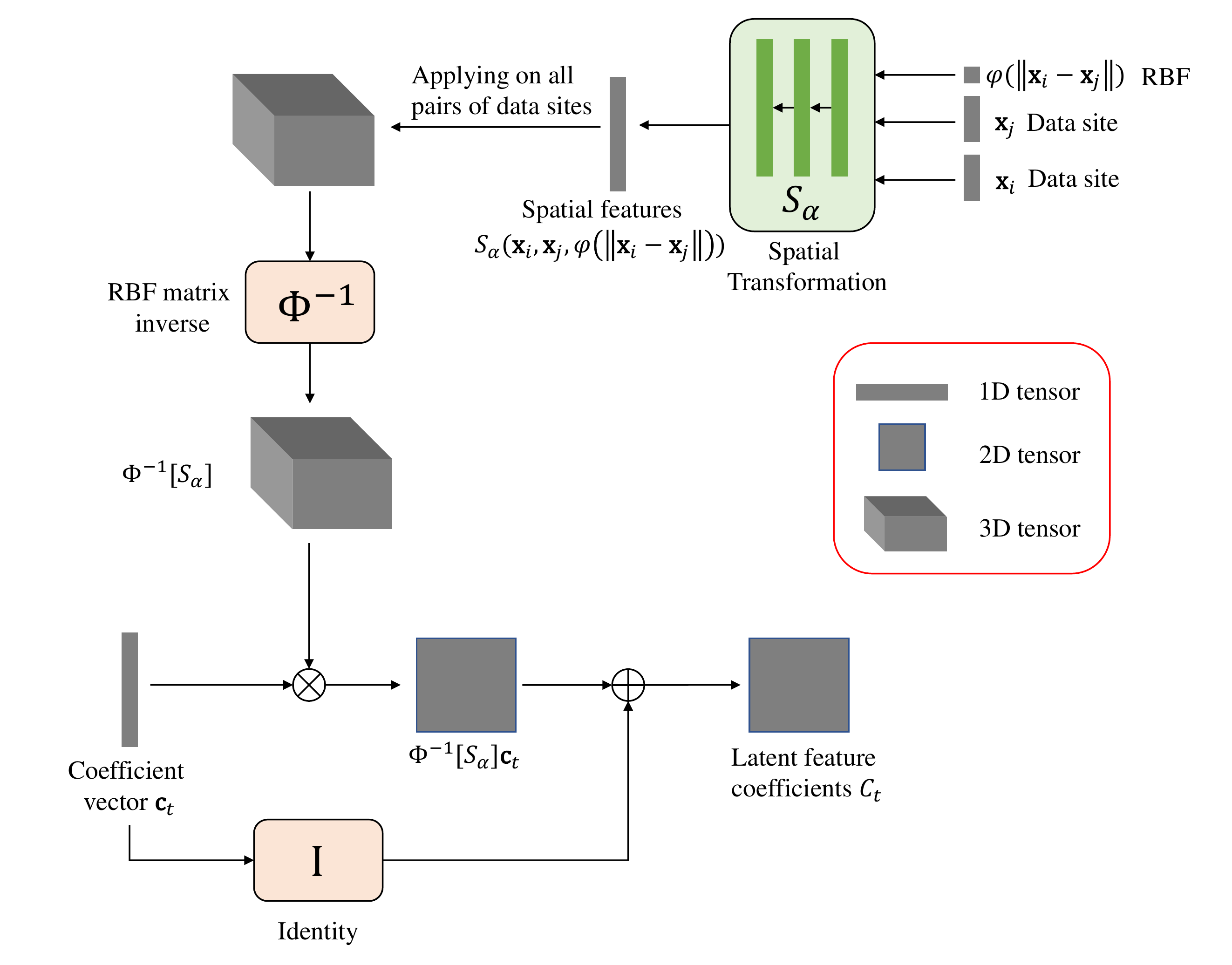}
\caption{A linear spatial transformation block. Latent spatial features are learned using a neural network $S_\alpha$, with parameter vector $\alpha$. The spatial features are multiplied with the RBF matrix inverse and RBF coefficient vector to generate latent feature coefficients with a residual connection.}
\label{fig:LSTB}
\end{figure}

For the next step of our method, we use transformed coefficient vector obtained by rewriting (\ref{eqn:compact_linear_dynamics}) as
\begin{equation}
    C_t = (I + \Phi^{-1} [S_\alpha])\mathbf{c}_t,
\end{equation}
where $I$ denotes the identity matrix. We generate multiple latent feature coefficient vectors (collectively denoted as $C_t$) by changing $S_\alpha$ to output more than one spatial features. The overall process of generating $C_t$ is termed as termed as \textit{Linear Spatial Transformation Block} (LSTB) and pictorially shown in Figure \ref{fig:LSTB}. LSTB  serves as a building block for the next step of our method, which aims to model a more generic nonlinear spatiotemporal dynamics. An LSTB provides some candidate linear (spatial) differential transformations present in the system. These linear (spatial) differential features are then used by two other types of neural network blocks which are responsible for uncovering any nonlinear relationship present in the system. We discuss these two types of  neural network blocks in the following two subsections.

\subsection{Recurrent Fusion Network (RFN)}
In \cite{saha2021deep}, we employed the spatial transformations only on the current measurement vector, whereas the past measurement vectors are only considered for approximating the temporal order of the underlying PDE. This assumption is valid only for spatiotemporal dynamical systems that can be purely modeled with a PDE of the considered form. However, many real-world dynamical systems entail complex spatiotemporal interaction over multiple time steps. Therefore, we consider applying the spatial transformations across multiple past measurement vectors and use a \textit{recurrent neural network} to unfold the relationships among them. We apply the LSTB on all available past measurement vectors $\mathbf{u}_{\leq t}$ 
to generate corresponding latent feature coefficient vectors $C_0^{(1)}, C_1^{(1)}, \cdots, C_t^{(1)}$. The superscript notation $\cdot^{(1)}$ will be apparent in the next subsection. The latent feature coefficient vectors $C_t^{(1)}$ are multiplied with the RBF matrix $\Phi$ to get the latent feature vectors $U_t^{(1)}$, i.e., 
\begin{equation}
    U_t^{(1)} = \Phi C_t^{(1)} 
\end{equation}
We finally feed these latent feature vectors $U_t^{(1)}$ to a recurrent network $R_\beta$, with parameter vector $\beta$, sequentially to get the next step prediction 
\begin{equation}
    \widehat{\mathbf{u}}_{t+1} = R_\beta \big(U_t^{(1)}, U_{t-1}^{(1)}, \cdots, U_0^{(1)} \big)
\end{equation}

\begin{figure}[t]
\centering\includegraphics[width=0.85\linewidth]{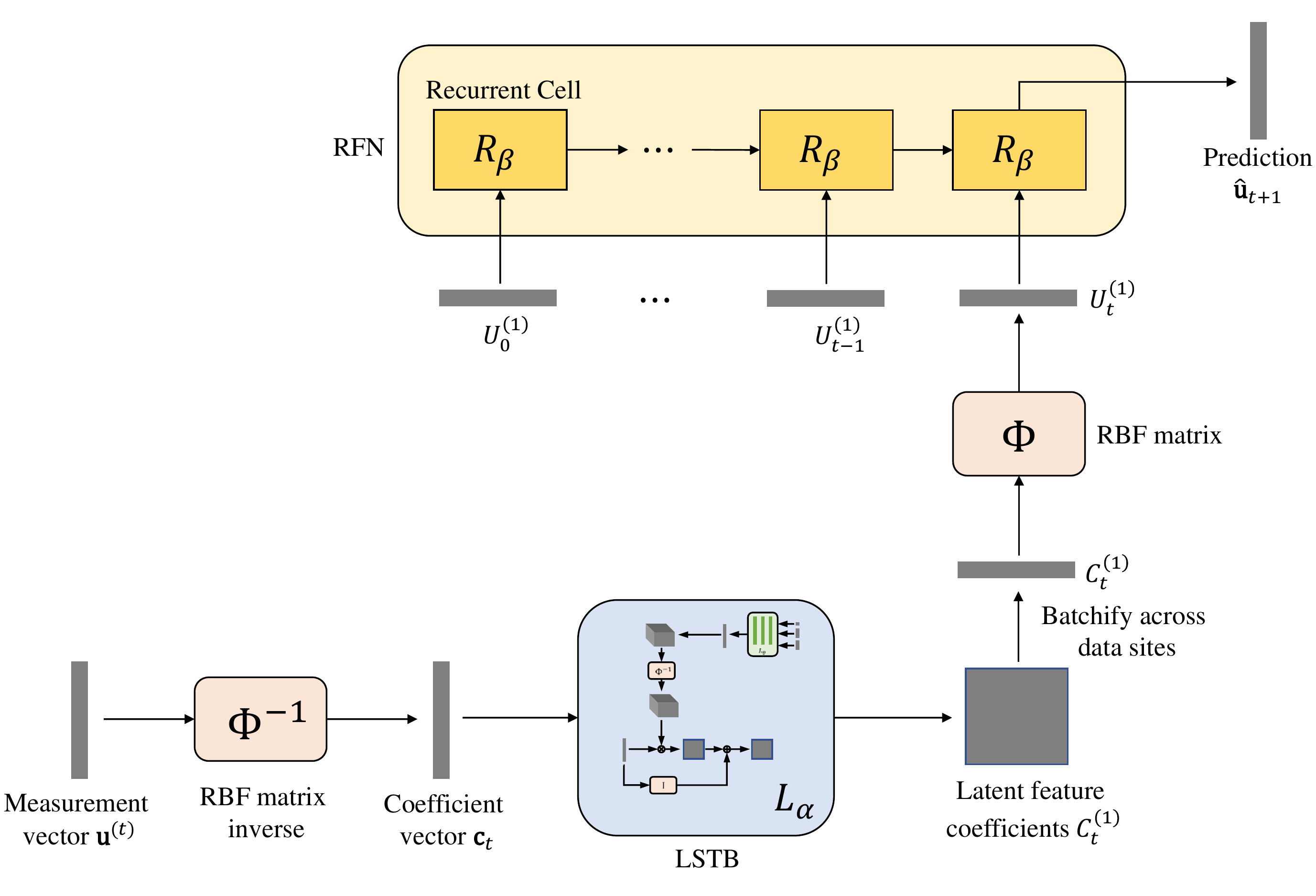}
\caption{End-to-end model involving a single LSTB $L_\alpha$, where $\alpha$ is the parameter vector of its spatial transformation neural network $S_\alpha$. $R_\beta$ denotes the recurrent cell with parameter vector $\beta$. Latent feature vectors from the past and current time steps are looped through the RFN to predict the future observations.}
\label{fig:onelevel}
\end{figure}

Figure \ref{fig:onelevel} shows the combined network involving an LSTB and a recurrent network. The name \textit{recurrent `fusion' network} (RFN) will become clear in the following subsection. 

\subsection{Nonlinear Aggregation Block (NAB) and Multilevel Transformation}
\begin{figure}[t]
\centering\includegraphics[width=0.85\linewidth]{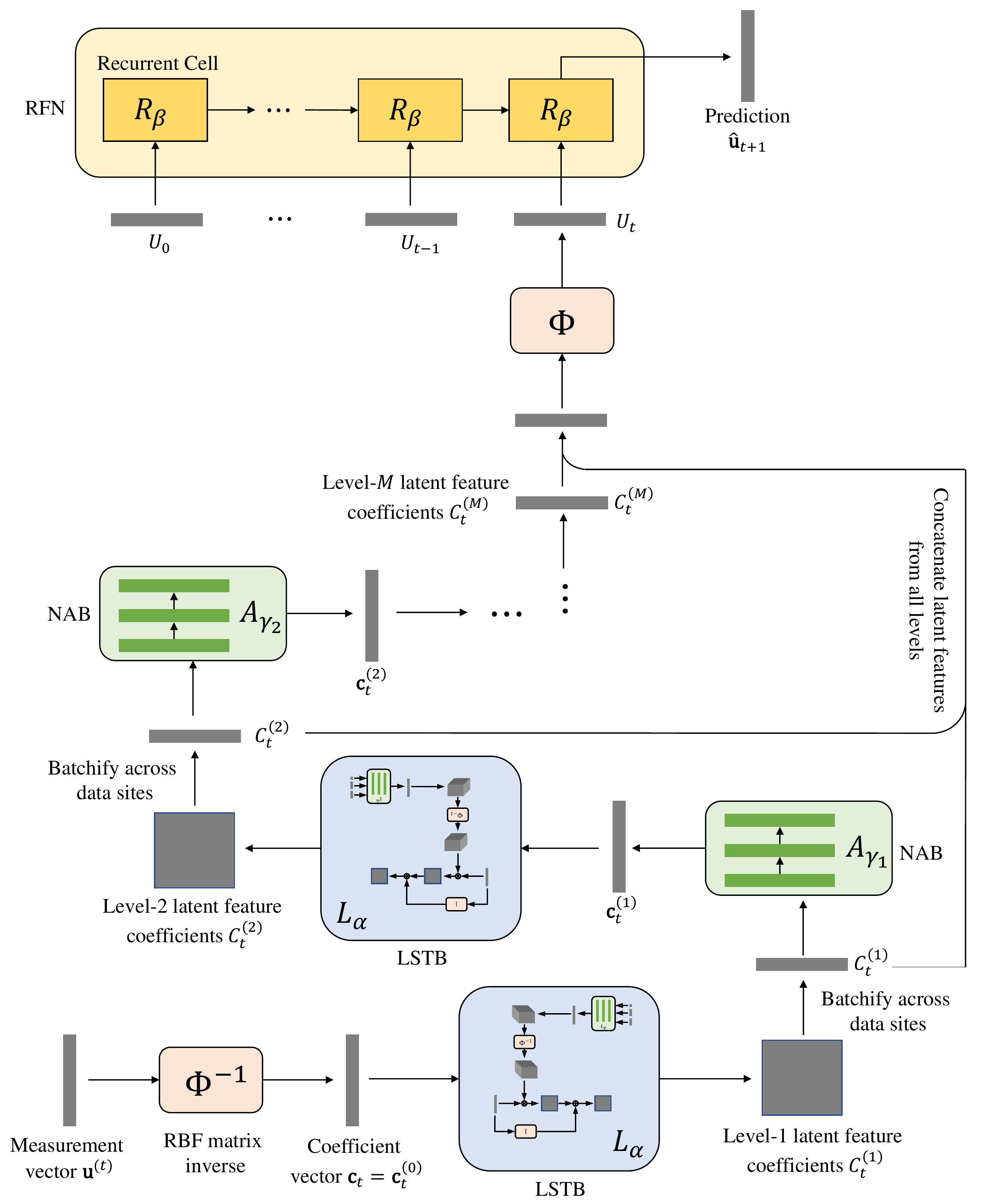}
\caption{End-to-end model involving multilevel transformations. Spatial transformation is applied to the nonlinear combination of latent feature coefficients at each level. NABs are feed-forward neural networks with parameter vectors $\gamma_m, m = 1,2, \cdots, M$.}
\label{fig:multilevel}
\end{figure}

A single LSTB-based model applies spatial transformation only on the coefficient vector $\mathbf{c}_t$ of the raw measurement vector $\mathbf{u}_t$. However, many dynamical systems involve spatial differential operations not only on the direct observation variable but also on its higher level nonlinear transformations. A single LSTB may not capture such complex operations accurately. Therefore, we propose to apply multilevel spatial transformation using a series of LSTBs. Consider the latent feature coefficient vectors after the first LSTB $C_t^{(1)}$, defined in the preceding two subsections, as \textit{level-1 latent feature coefficient vectors}. Note that $C_t^{(1)}$ comprises coefficients corresponding to multiple spatial features, as mentioned in section \ref{method}\ref{lstb}. We use a feed-forward neural network $A_{\gamma_1}$, with parameter vector $\gamma_1$, to generate a single nonlinear combination feature coefficient vector $\mathbf{c}_t^{(1)}$, i.e.,
\begin{equation}
    \mathbf{c}_t^{(1)} = A_{\gamma_1}\big(C_t^{(1)}\big)
\end{equation}
Output $\mathbf{c}_t^{(1)}$ from the \textit{nonlinear aggregation block} (NAB) $A_{\gamma_1}$ is then fed to another LSTB to generate level-2 latent feature coefficient vectors $C_t^{(2)}$. We repeat the process for a cascade of LSTB-NAB pairs to generate a set of multilevel latent feature coefficient vectors $C_t^{(m)}, m = 1, 2, \cdots, M$. Latent feature coefficient vectors from different levels are concatenated and then multiplied with RBF matrix $\Phi$ to get the latent feature vectors $U_t$, i.e., 
\begin{equation}
    U_t = \Phi \big[\mathbf{c}_t^{(0)} \ | \  C_t^{(1)} \ | \ \cdots \ | \ C_t^{(M)}\big]
    \label{eqn:concat}
\end{equation}
Here $|$ denotes the concatenation operation. Note, we have included the 0-level feature coefficient vector $\mathbf{c}_t^{(0)}$ as well in (\ref{eqn:concat}) which is basically the coefficient vector corresponding to the raw measurement vector, i.e., $\mathbf{c}_t^{(0)} = \mathbf{c}_t$. Similar to the one-level method described in the preceding subsection, we generate latent feature vectors $U_t$ 
by applying the cascade of LSTB-NAB pairs on all available past measurement vectors $\mathbf{u}_{\leq t}$. 
These multilevel feature vectors $U_t$  
are finally `fused' sequentially in a recurrent fusion network to get the next step prediction 
\begin{equation}
    \widehat{\mathbf{u}}_{t+1} = R_\beta (U_t, U_{t-1}, \cdots, U_0 )
\end{equation}

The overall process involving multiple LSTBs, NABs and a RFN is depicted in Figure \ref{fig:multilevel}. 

\paragraph{Learning Objective}
We train the proposed unraveled multilevel transformation network for multi-step predictions by optimizing the following objective:
\begin{equation}
    \min_{\theta := [\alpha, \beta, \gamma_1, \gamma_2, \cdots, \gamma_M]} 
    \sum_{t=1}^{T+\tau-1} \| \mathbf{u}_t - \widehat{\mathbf{u}}_t \|^2
\end{equation}

%% file: expSetup.tex
\section{Experiment Setup}

\subsection{Model configuration}
\paragraph{RBF} We use \textit{multiquadric} $\varphi(\|\mathbf{x}\|) = \sqrt{\|\mathbf{x}\|^2 + \epsilon^2}$ as the RBF, where the shape parameter $\epsilon$ adjusts the flatness/steepness of the function. The RBF and its shape parameter are chosen using leave-one-out cross-validation on the training dataset, which is common in RBF-based interpolation methods. Value at one randomly selected left-out node for each training sample is estimated using the values at other nodes by means of RBF interpolation. We compute the average leave-one-out cross-validation (estimation) error on the training dataset for different RBFs, e.g., multiquadric, inverse multiquadric, Gaussian, thin-plate splines etc., and different sets of shape parameter values. We choose multiquadric RBF because it showed lowest error for all datasets.  However, the shape parameter $\epsilon$ is different for different datasets.

\paragraph{Spatial transformation $S_\alpha$} For the spatial transformation network inside an LSTB, we use a \textit{multilayer perceptron} (MLP) with two hidden layers of sizes $64$ and $32$, respectively, with ReLU activation. This network takes a vector of size $(2d+1)$ as input, $d$ being the dimension of the spatial domain, and outputs a spatial feature vector of size $8$. The parameter vector $\alpha$ is shared across all LSTBs as well as across all pairs of data sites. 

\paragraph{NAB $A_\gamma$} Nonlinear aggregation blocks are implemented using fully-connected networks with one hidden layer of size $32$ with ReLU activation. Input layer and output layer dimensions are $8$ and $1$, respectively. The parameter vectors $\gamma_m, m = 1, 2, \cdots, M$ of the NABs are shared across all data sites.

\paragraph{RFN $R_\beta$} A \textit{Gated Recurrent Unit} (GRU) network with a single hidden layer of size $64$ is used for the RFN. Hidden cell output is passed through a linear layer to provide the final output of size $1$. Input size of the RFN is $(8M+1)$, where $M$ denotes the number of levels. 
The parameter vector $\beta$ is shared across all data sites.

\subsection{Baseline models}
We compare the performance of our model against the following two baselines that are designed to learn from scattered data.
\paragraph{DRC \cite{saha2021deep}} \textit{Deep RBF Collocation} (DRC) is a cascade of a spatial transformation block  and a neural aggregator. The spatial transformation block is an MLP with two hidden layers of sizes $64$ and $32$, respectively, with ReLU activations. This network takes a vector of size $(2d+1)$ as input, $d$ being the dimension of the spatial domain, and outputs a spatial feature vector of size $16$. The neural aggregator is also an MLP with three hidden layers of sizes 128, 64, and 32, respectively, with ReLU activations.
\paragraph{PADGN \cite{seo2019physics}}  
\textit{Physics-aware Difference Graph Networks} is a graph-based neural network with physics-aware spatial difference layer. 
The spatial derivative layer is message passing neural network with two graph network (GN) blocks involving 2-layer MLPs. The forward network is a recurrent graph neural network with two recurrent GN blocks involving 2-layer GRU cells. We use the original model configuration mentioned in \cite{seo2019physics}. 

Table \ref{tab:param_count} compares the total number of learnable parameters of our model UMTN and the baselines.

\begin{table}[!h]
\caption{Number of parameters for different models}
\label{tab:param_count}
\begin{tabular*}{\textwidth}{c @{\extracolsep{\fill}} ccc}
\toprule
\textbf{Model} &\textbf{PADGN}\cite{seo2019physics} &\textbf{DRC}\cite{saha2021deep}  &\textbf{UMTN (3-level)} \\
\midrule
parameter-count   &$340,001$ &$15,474$ & $20,586$\\ 
\bottomrule
\end{tabular*}
\vspace*{-4pt}
\end{table}

\subsection{Training settings}
We train our model using Adam optimizer with learning rate of $0.001$ for a maximum of $1000$ epochs with early stopping. Batch size chosen according to total available examples in each dataset. We use inverse sigmoid scheduled sampling \cite{bengio2015scheduled}  during training with the coefficient $k = 50$. The shape parameter $\epsilon$ of the RBF is chosen using \textit{leave-one-out cross-validation} method \cite{rippa1999algorithm,fasshauer2007choosing} on the training set. The coefficient vector $\mathbf{c}_t$ corresponding to the raw measurement vector $\mathbf{u}_t$ is obtained by solving (\ref{eqn:compact_linear_sys}) using $\ell_2$-regularized least square, with a regularization parameter $= 10^{-2}$, instead of direct inversion since $\Phi^{-1}$ can have large condition number depending on the data sites distribution. For the same reason, $\Phi^{-1}$ inside the LSTB is scaled to have values in range $[-1,1]$.

Same training settings are used for the DRC baseline, whereas for PADGN, we use the training settings of \cite{seo2019physics}.

\subsection{Evaluation Metric}
For all experiments, data values are normalized using the mean and variance of the corresponding training set to have zero mean and unit variance. To evaluate prediction accuracy, we use the Mean Absolute Error (MAE, lower is better) between the normalized ground truth and model prediction over all data sites, defined as follows:
\begin{equation}
    \text{MAE} = \frac{1}{Tn} \sum_{t=\tau}^{T+\tau-1} \sum_{j=1}^n |u_{\mathcal{N}}(t, \mathbf{x}_j) - \widehat{u}_{\mathcal{N}}(t, \mathbf{x}_j)|, 
\end{equation}
where $u_{\mathcal{N}}$ and $\widehat{u}_{\mathcal{N}}$ denote the normalized ground truth and prediction, respectively. 
The computed error does not depend on the absolute values of the measured/observed data which vary across different datasets.
We repeat each experiment (both training and evaluation) 3 times and report the average and standard deviation of MAE.

%% file: results.tex
\section{Results}
We evaluate our models and baselines on both synthetic (or simulated) dataset and real-world datasets. For synthetic dataset, we use \textit{convection-diffusion equation}. The two real-world datasets contains spatiotemporal observations of \textit{atmospheric temperature} and \textit{sea-surface temperature} (SST), respectively. All the datasets are collected from the public repository of \cite{seo2019physics}. For our model UMTN, we report results up to 3 levels. UMTN 1-level shows the advantage of employing recurrent network instead of feed-forward network as used in DRC \cite{saha2021deep}. The effect of adding more levels is demonstrated using UMTN 2-level and UMTN 3-level.  

\subsection{Convection-Diffusion Equation}

\paragraph{Dataset description} The convection-–diffusion equation describes the flow of energy, particles, or other physical quantities inside a physical system involving both diffusion and convection. We consider the following linear variable--coefficient convection--diffusion equation as used by the authors in \cite{seo2019physics}, 
\begin{equation}
    \frac{\partial u}{\partial t} = [a(\mathbf{x}), b(\mathbf{x})]^\top \nabla u  + c(\mathbf{x}) \nabla^2 u, \quad \mathbf{x} \in \Omega \subset \mathbb{R}^2, \quad 0 \leq t \leq 0.2
\end{equation}
with initial condition 
\begin{equation}
    u(0, \mathbf{x}) = \sum_{|k|,|l| \leq 9} \lambda_{k,l} \cos([k, l]^\top \mathbf{x}) + \zeta_{k,l} \sin([k, l]^\top \mathbf{x})
    \label{eqn:init_cond}
\end{equation}
Here, $\mathbf{x} = [x, y]^\top$, \  $\Omega = [0, 2\pi] \times [0, 2\pi]$, \  $a(\mathbf{x}) = a(x, y) = 0.5 (\cos(y) + x(2\pi - x) \sin(x)) + 0.6$, \ $b(\mathbf{x}) = b(x, y) = 2 (\cos(y) + \sin(x)) + 0.8$, \  and \ $c(\mathbf{x}) = c(x,y) = 0.5\big(1 - \frac{1}{\sqrt{2}\pi}\sqrt{(x-\pi)^2 + (y-\pi)^2}\big)$. $\nabla$ and $\nabla^2$ denote the gradient operator and Laplace operator, respectively.  $\lambda_{k,l} ,  \zeta_{k,l}$ are sampled from the normal distribution $\mathcal{N}(0, 0.02)$ and $k,l$ are chosen randomly. 

The dataset contains $1000$ time series of $u$ at $250$ data sites uniformly sampled from snapshots that are generated on a $50 \times 50$ regular mesh with time step-size $\Delta t = 0.01$ and random initial condition defined in (\ref{eqn:init_cond}). We use $700$ time series for training, $150$ for validation, and $150$ for test.

\paragraph{Prediction performance} We evaluate the methods on the tasks of predicting data (measurement) values at all the data sites for $T$ future time steps given observed values of the first $\tau$ time steps. For this experiment, we choose $\tau=5$ and $T=15$ as in \cite{seo2019physics}.

\begin{figure}[h]
\centering\includegraphics[width=1\linewidth]{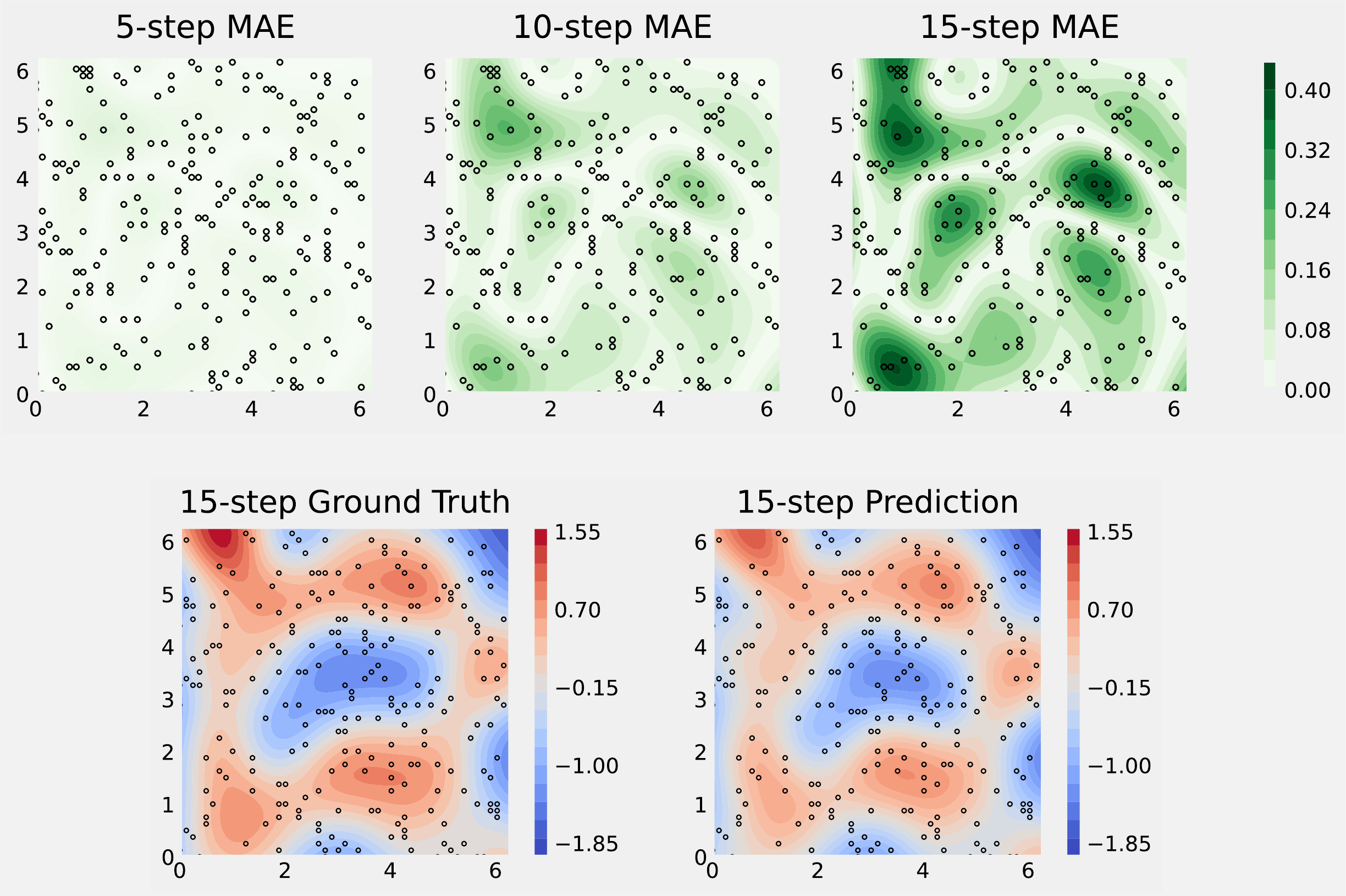}
\caption{Top row: Interpolated MAE distribution for our method over time for our method in the convection-diffusion experiment. All three MAE figures share same colorbar shown in the right.  
Bottom row: Interpolated 15$^{th}$ step ground truth and prediction. The circles denote the data sites.}
\label{fig:synthetic}
\end{figure}

Table \ref{tab:mae_synthetic} compares the prediction performance of different models in terms of mean absolute error. Our proposed model with $3$-level transformation (cascade of 3 LSTB-NAB pairs) outperforms the baselines. Increasing the number of levels up to 3 improves the accuracy. Incorporating more levels contributes to negligible improvement in accuracy. The effect of the RFN can be observed by comparing DRC \cite{saha2021deep} and UMTN 1-level.
An example of 15-step (interpolated) ground truth and prediction is shown in Figure \ref{fig:synthetic} (bottom row). MAE distribution across data sites, at different time steps, for the same example is shown as well (Figure \ref{fig:synthetic}, top row).  

\begin{table}[!h]
\caption{15-step mean absolute error for the convection-diffusion experiment}
\label{tab:mae_synthetic}
\begin{tabular*}{\textwidth}{c @{\extracolsep{\fill}} ccccc}
\toprule
\multirow{2}{*}{\textbf{Model}}   &\multirow{2}{*}{\textbf{PADGN}\cite{seo2019physics}} &\multirow{2}{*}{\textbf{DRC}\cite{saha2021deep}} &\multicolumn{3}{c}{\textbf{UMTN}} \\
& & & 1-level & 2-level & 3-level\\
\midrule
\multirow{2}{*}{MAE} &$0.1087$ & $0.1726$ & $0.1343$ & $0.1093$ & $\mathbf{0.1070}$ \\
 & $\pm 0.0098$ & $\pm 0.0046$ & $\pm 0.0138$ & $\pm 0.0043$ & $\pm 0.0021$ \\
\bottomrule
\end{tabular*}
\end{table}

\subsection{NOAA Atmospheric Temperature Dataset}

\paragraph{Dataset description} This dataset contains meteorological observations (temperature) at the land-based weather stations located in the United States, collected from the Online Climate Data Directory of the National Oceanic and Atmospheric Administration (NOAA). The weather stations are sampled from the \textit{Western} and \textit{Southeastern} states that have actively measured meteorological observations during 2015 \cite{seo2019physics}. The 1-year sequential data of hourly temperature record are divided into small sequences of 24 hours. For training, validation and test a sequential 8-2-2 (months) split is used.  

\paragraph{Prediction performance} We evaluate the methods on the tasks of predicting data (measurement) values at all the data sites for $T$ future time steps given observed values of the first $\tau$ time steps. We choose $\tau=12$ as in \cite{seo2019physics} and investigate both single-step ($T=1$) and multi-step ($T=6$ and $T=12$) predictions.

\begin{figure}[t]
\centering\includegraphics[width=0.8\linewidth]{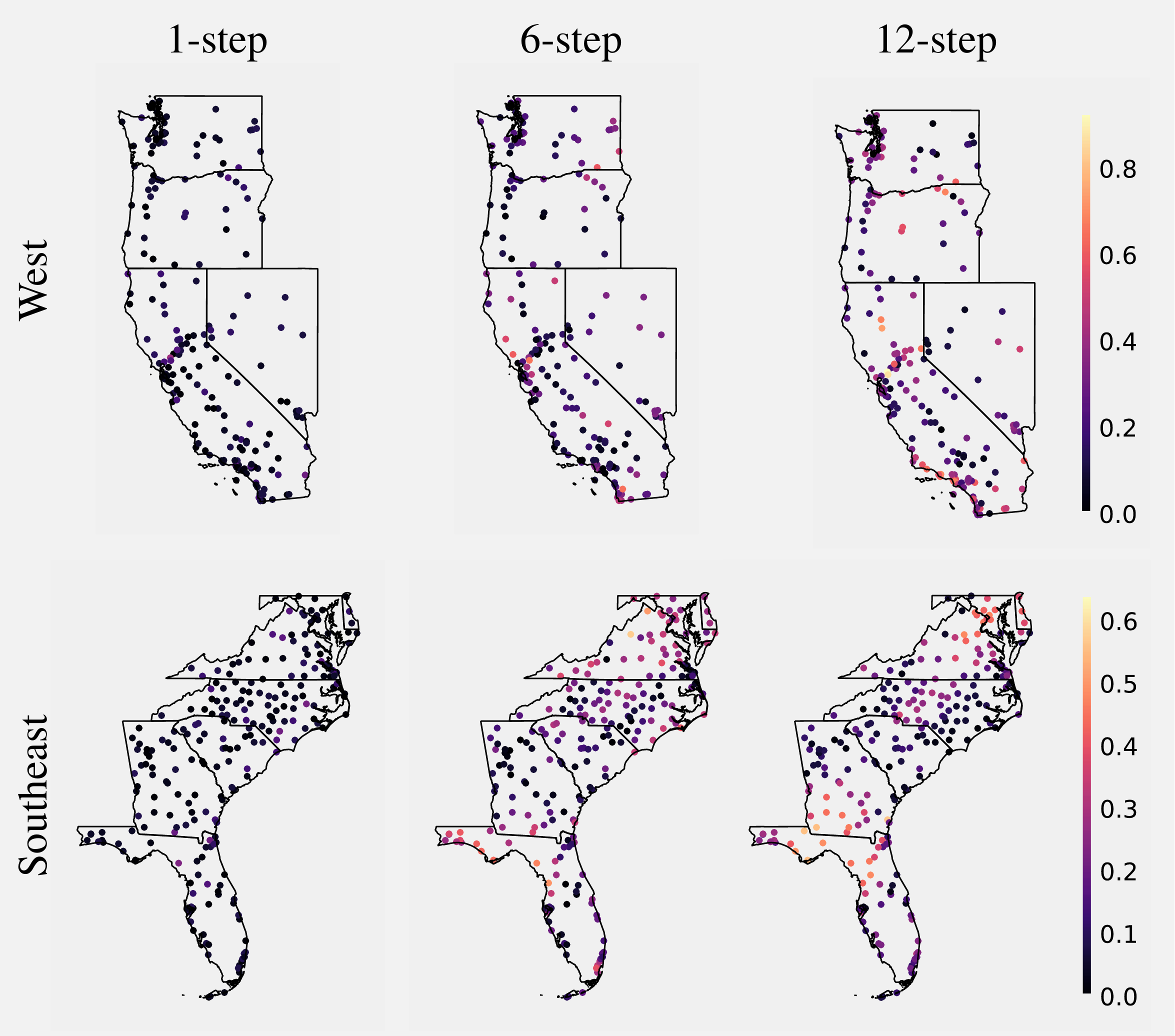}
\caption{MAE distribution over time across data sites for our method in the NOAA experiment. All figures in each row share the same colorbar, shown in the right of the corresponding row.}
\label{fig:noaa}
\end{figure}

Prediction performance of different models in terms of mean absolute error is shown in Table \ref{tab:mae_noaa}. Our 3-level model outperforms the baselines in multi-step prediction (6-step and 12-step) for both the regions. Adding the RFN contributes to a large drop in MAE from DRC \cite{saha2021deep} to UMTN 1-level. It implies that the temporal relation across past observations are more complex for real datasets.   
An example of MAE distribution across data sites, at different time steps, is shown in Figure \ref{fig:noaa}.

\begin{table}[!h]
\centering
\caption{Mean absolute error for the NOAA experiment}
\label{tab:mae_noaa}
\begin{tabular*}{\textwidth}{c @{\extracolsep{\fill}} ccccc}
\toprule
\textbf{Region} & \textbf{Model} & &\textbf{1-step} &\textbf{6-step} &\textbf{12-step} \\
\midrule
\multirow{5}{*}{West} & PADGN & & $\mathbf{0.0840} \pm 0.0004$ & $0.1614 \pm 0.0042$ & $0.2439 \pm 0.0163$  \\ 
& DRC & & $0.1655 \pm 0.0087$ & $0.3333 \pm 0.0409$ & $0.4378 \pm 0.0641$ \\
& \multirow{3}{*}{UMTN} & 1-level & $0.0894 \pm 0.0025$ & $0.1702 \pm 0.0147$ & $0.2311 \pm 0.0198$ \\
& & 2-level & $0.0927 \pm 0.0022$ & $0.1711 \pm 0.0042$ & $0.2361 \pm 0.0061$\\
& & 3-level & $0.0861 \pm 0.0020$ & $\mathbf{0.1539} \pm 0.0062$ & $\mathbf{0.2090} \pm 0.0116$\\

\midrule
\multirow{5}{*}{SouthEast} & PADGN &  & $0.0721 \pm 0.0002$ &  $0.1664 \pm 0.0011$  & $0.2408 \pm 0.0056$\\
& DRC & & $0.1450 \pm 0.0048$ &  $0.3052 \pm 0.0197$  & $0.3824 \pm 0.0262$\\
& \multirow{3}{*}{UMTN} & 1-level & $\mathbf{0.0652} \pm 0.0018$ & $0.1553 \pm 0.0179$ & $0.2110 \pm 0.0165$ \\
& & 2-level & $0.0674 \pm 0.0016$ & $0.1677 \pm 0.0055$ & $0.2070 \pm 0.0099$\\
& & 3-level & $0.0661 \pm 0.0007$ & $\mathbf{0.1535} \pm 0.0081$ & $\mathbf{0.2049} \pm 0.0064$ \\
\bottomrule
\end{tabular*}
\vspace*{-4pt}
\end{table}

\subsection{NEMO Sea Surface Temperature Dataset}

\paragraph{Dataset description} This dataset contains saptiotempral sequences of SST generated by the NEMO ocean engine \cite{madec2017nemo}.\footnote{ Available at \href{http://marine.copernicus.eu/services-portfolio/access-to-products/?option=com_csw&view=
details&product_id=GLOBAL_ANALYSIS_FORECAST_PHY_001_024}{http://marine.copernicus.eu/services-portfolio/access-to-products/?option=com\_csw\&view=details\&product\\\_id=GLOBAL\_ANALYSIS\_FORECAST\_PHY\_001\_024} }. The observations correspond to $250$ randomly selected data sites within a $[0,550] \times [100,650]$ square cropped from the area between $50\text{N}\degree  \-- 65\text{N}\degree$ and $75\text{W}\degree \-- 10\text{W}\degree$
starting from 01-01-2016 to 12-31-2017 \cite{seo2019physics}. The data is divided into 24 sequences, each lasting 30 days (extra days in each month are truncated). Data corresponding to 2016 are used for training and the rest is used for validation and testing, in equal sequential split.

\paragraph{Prediction performance} We evaluate the methods on the tasks of predicting data (measurement) values at all the data sites for $T$ future time steps given observed values of the first $\tau$ time steps. We choose $\tau=5$ as in \cite{seo2019physics} and investigate two cases: $T=15$ and $T=25$.

\begin{figure}[h]
\centering\includegraphics[width=1\linewidth]{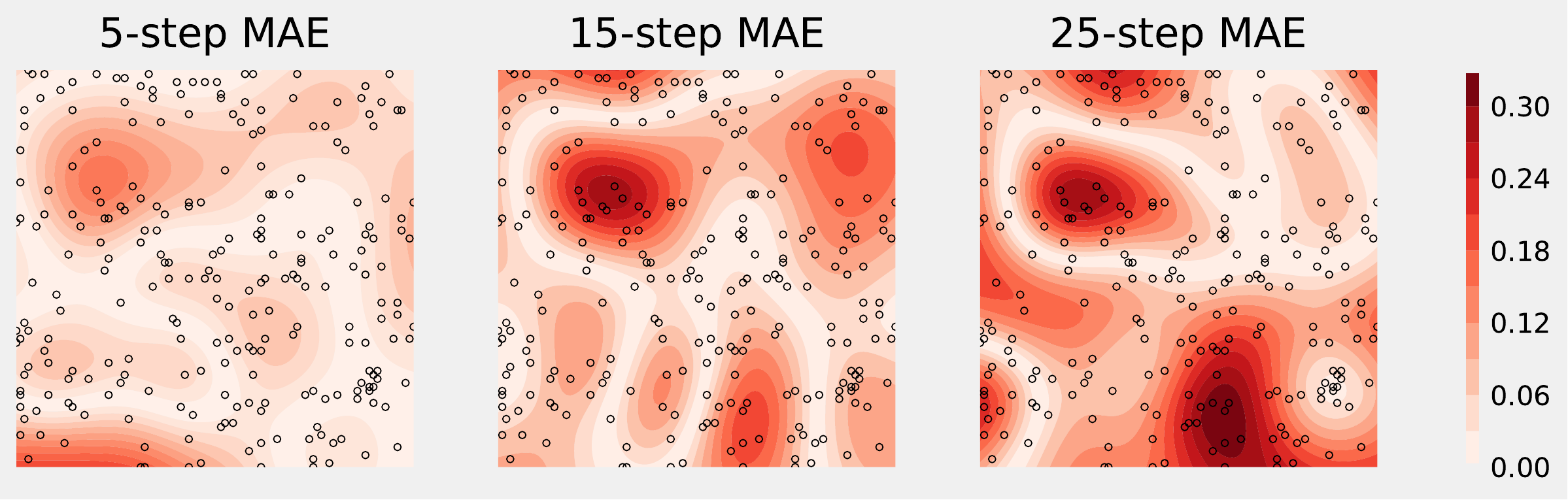}
\caption{Interpolated MAE distribution over time across data sites for our method in the SST experiment. All three figures share same colorbar shown in the right. The circles denote the data sites.}
\label{fig:sst}
\end{figure}

Table \ref{tab:mae_sst} compares the prediction accuracy of different models in terms of mean absolute error. Our 3-level model outperforms the baselines in 25-step prediction. Drop in MAE from 1-level UMTN to 2-level UMTN is relatively larger than the drop in MAE from 2-level UMTN to 3-level and adding more levels contributes to negligible improvement in accuracy. 
An example of MAE distribution across data sites, at different time steps, is shown in Figure \ref{fig:sst}.

\begin{table}[!h]
\centering
\caption{Mean absolute error for the SST experiment}
\label{tab:mae_sst}
\begin{tabular*}{0.7\textwidth}{c @{\extracolsep{\fill}} ccc}
\toprule
\textbf{Model} & &\textbf{15-step} &\textbf{25-step} \\
\midrule
PADGN & & $0.1437 \pm 0.0007$ & $0.1850 \pm 0.0015$  \\ 
DRC & & $0.1690 \pm 0.0087$ & $0.2058 \pm 0.0103$ \\
\multirow{3}{*}{UMTN} & 1-level &  $0.1526 \pm 0.0009$ & $0.1958 \pm 0.0005$\\
& 2-level &  $0.1440 \pm 0.0003$ & $0.1820 \pm 0.0003$ \\
& 3-level &  $0.1436 \pm 0.0006$ & $\mathbf{0.1812} \pm 0.0009$ \\
\bottomrule
\end{tabular*}
\vspace*{-4pt}
\end{table}

%% file: conclusion.tex
\section{Conclusion}
We have introduced a framework for data-driven prediction of spatiotemporal dynamics when data sites are sparse and irregularly distributed. The proposed method does not assume any specific physical representation of the underlying dynamical system and is applicable to any spatiotemporal dynamical systems involving continuous state variables. We demonstrated superior predictive performance of our model using both simulated and real datasets. The proposed method can be straightforwardly adapted to system involving multiple variables by applying the LSTB to each of the measurement variables to create their corresponding linear (spatial) differential features. NAB and RFN can then be used to learn the nonlinear relationships among those linear (spatial) differential features of different measurement variables.       

The current method only consider spatial irregularity and sparsity but assumes regular sampling in time. Augmenting the current architecture with models like latent ODEs or ODE-RNNs \cite{chen2018neural, rubanova2019latent} for irregular time samples would be an important extension of the current work. Since our method provides a framework for prediction from sparsely-observed data, a natural future direction would be to incorporate uncertainty \cite{gal2016uncertainty} in prediction due to lack of information.    

Predicting real-world spatiotemporal processes purely from data is an extremely challenging problem. Though our method shows promising performance on data-driven prediction spatiotemporal dynamics without any prior knowledge about the physical system, like any other pure data-driven model, it is unlikely to provide best results in all environments. Therefore, rather than considering the proposed model as a standalone, as a future work, we should reinforce it with physical models, following the recent line of work \cite{long2018hybridnet, saha2021deep, guen2020augmenting}, for more accurate and reliable prediction.    